\documentclass[11pt,a4paper]{article}
\usepackage[hyperref]{emnlp2018}
\usepackage{times}
\usepackage{latexsym}
\usepackage{graphicx}
\usepackage{multicol}
\usepackage{amsmath,mathtools}
\usepackage{todonotes}
\usepackage{url}

\aclfinalcopy 

\newcommand\blfootnote[1]{%
  \begingroup
  \renewcommand\thefootnote{}\footnote{#1}%
  \addtocounter{footnote}{-1}%
  \endgroup
}

\title{DeFactoNLP: Fact Verification using Entity Recognition, TFIDF Vector Comparison and Decomposable Attention}

\author{Aniketh Janardhan Reddy\textsuperscript{*}\\
        Machine Learning Department \\
        Carnegie Mellon University\\
        Pittsburgh, USA \\
        {\tt ajreddy@cs.cmu.edu}\\\And
        Gil Rocha \\
        LIACC/DEI \\
        Faculdade de Engenharia \\
        Universidade do Porto, Portugal \\
        {\tt gil.rocha@fe.up.pt} \\\And
        Diego Esteves \\
        SDA Research \\
        University of Bonn \\ Bonn, Germany \\
        {\tt esteves@cs.uni-bonn.de}}

\date{}

\begin{document}
\maketitle
\begin{abstract}
\blfootnote{\textsuperscript{*}Work was completed while the author was a student at the Birla Institute of Technology and Science, India and was interning at SDA Research.}
In this paper, we describe DeFactoNLP\footnote{\url{https://github.com/DeFacto/DeFactoNLP}}, the system we designed for the FEVER 2018 Shared Task. The aim of this task was to conceive a system that can not only automatically assess the veracity of a claim but also retrieve evidence supporting this assessment from Wikipedia. 
In our approach, the Wikipedia documents whose Term Frequency-Inverse Document Frequency (TFIDF) vectors are most similar to the vector of the claim and those documents whose names are similar to those of the named entities (NEs) mentioned in the claim are identified as the documents which might contain evidence. 
The sentences in these documents are then supplied to a textual entailment recognition module. This module calculates the probability of each sentence supporting the claim, contradicting the claim or not providing any relevant information to assess the veracity of the claim. Various features computed using these probabilities are finally used by a Random Forest classifier to determine the overall truthfulness of the claim. The sentences which support this classification are returned as evidence. Our approach achieved\footnote{The scores and ranks reported in this paper are provisional and were determined prior to any human evaluation of those evidences that were retrieved by the proposed systems but were not identified in the previous rounds of annotation. The organizers of the task plan to update these results after an additional round of annotation.} a 0.4277 evidence F1-score, a 0.5136 label accuracy and a 0.3833 FEVER score\footnote{FEVER score measures the fraction of claims for which at least one complete set of evidences have been retrieved by the fact verification system.}.
\end{abstract}

\section{Introduction}

Given the current trend of massive fake news propagation on social media, the world is desperately in need of automated fact checking systems. Automatically determining the authenticity of a fact is a challenging task that requires the collection and assimilation of a large amount of information. To perform the task, a system is required to find relevant documents, detect and label evidences, and finally output a score which represents the truthfulness of the given claim. The numerous design challenges associated with such systems are discussed by \citet{Thorne2018AutomatedFC} and \citet{esteves2018toward}.


The Fact Extraction and Verification (FEVER) dataset~\cite{Thorne18Fever} is the first publicly available large-scale dataset designed to facilitate the training and testing of automated fact verification systems. The FEVER 2018 Shared Task required us to design such systems using this dataset. The organizers had provided us a preprocessed version of the June 2017 Wikipedia dump in which the pages only contained the introductory sections of the respective Wikipedia pages. Given a claim, we were asked to build systems which could determine 
if there were sentences supporting the claim (labelled as "SUPPORTS") or sentences refuting it (labelled as "REFUTES"). If conclusive evidence either supporting or refuting the claim could not be found in the dump, the system should report the same (labelled "NOT ENOUGH INFO"). However, if conclusive evidence was found, it should also retrieve the sentences which either support or refute the claim.
\section{System Architecture}
\label{sec:overview}
\begin{figure} 
  \footnotesize
  \centering
  \includegraphics[scale=0.4]{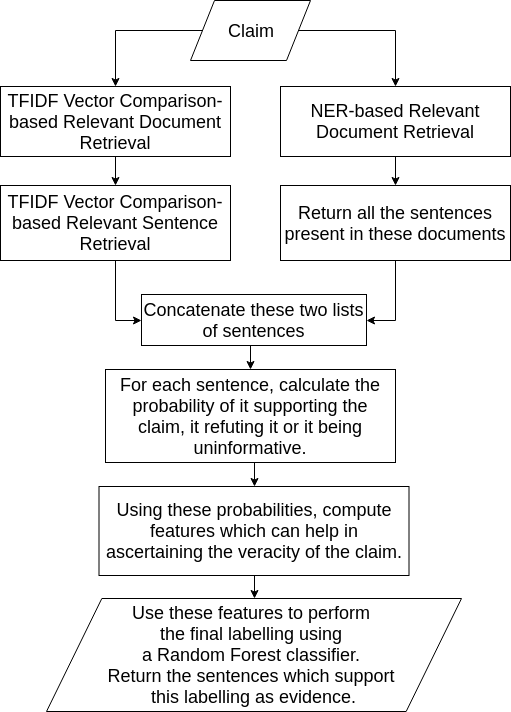}
  \caption{The main steps of our approach}
  \label{fig}
\end{figure}

Our approach has four main steps: Relevant Document Retrieval, Relevant Sentence Retrieval, Textual Entailment Recognition and Final Scoring and Classification. Given a claim, Named Entity Recognition (NER)  and TFIDF vector comparision are first used to retrieve the relevant documents and sentences as delineated in Section~\ref{sec:docsent_retrieval}. The relevant sentences are then supplied to the textual entailment recognition module (Section~\ref{sec:rte}) that returns a set of probabilities. Finally, a Random Forest classifier \cite{Breiman:2001:RF:570181.570182} is employed to assign a label to the claim using certain features derived from the probabilities returned by the entailment model as detailed in Section~\ref{sec:final}. The proposed architecture is depicted in Figure~\ref{fig}.

\subsection{Retrieval of Relevant Documents and Sentences}
\label{sec:docsent_retrieval}
We used two methods to identify which Wikipedia documents may contain relevant evidences. Information about the NEs mentioned in a claim can be helpful in determining the claim's veracity. In order to get the Wikipedia documents which describe them, the first method initially uses the Conditional Random Fields-based Stanford NER software~\cite{stanfordner} to recognize the NEs mentioned in the claim. Then, for every NE which is recognized, it finds the document whose name has the least Levenshtein distance~\cite{1966SPhD...10..707L} to that of the NE. 
Hence, we obtain a set of documents which contain information about the NEs mentioned in a claim. Since all of the sentences in such documents might aid the verification, they are all returned as possible evidences.

The second method used to retrieve candidate evidences is identical to that used in the baseline system~\cite{Thorne18Fever} and is based on the rationale that sentences which contain terms similar to those present in the claim are likely to help the verification process. Directly evaluating all of the sentences in the dump is computationally expensive. Hence, the system first retrieves the five most similar documents based on the cosine similarity between binned unigram and bigram TFIDF vectors of the documents and the claim using the DrQA system~\cite{chen2017reading}. Of all the sentences present in these documents, the five most similar sentences based on the cosine similarity between the binned bigram TFIDF vectors of the sentences and the claim are finally chosen as possible sources of evidence. The number of documents and sentences chosen is based on the analysis presented in the aforementioned work by \citet{Thorne18Fever}. 

The sets of sentences returned by the two methods are combined and fed to the textual entailment recognition module described in Section~\ref{sec:rte}.

\subsection{Textual Entailment Recognition Module}
\label{sec:rte}
Recognizing Textual Entailment (RTE) is the process of determining whether a text fragment (Hypothesis $H$) can be inferred from another fragment (Text $T$) \cite{SVR12}.
The RTE module receives the claim and the set of possible evidential sentences from the previous step. Let there be $n$ possible sources of evidence for verifying a claim. For the $i^{th}$ possible evidence, let $s_i$ denote the probability of it entailing the claim, let $r_i$ denote the probability of it contradicting the claim, and let $u_i$ be the probability of it being uninformative. The RTE module calculates each of these probabilities.  

The SNLI corpus \cite{BAPM15} is used for training the RTE model. This corpus is composed of sentence pairs $\langle T,H \rangle$, where $T$ corresponds to the literal description of an image and $H$ is a manually created sentence. If $H$ can be inferred from $T$, the ``Entailment'' label is assigned to the pair. If $H$ contradicts the information in $T$, the pair is labelled as ``Contradiction''. 
Otherwise, the label ``Neutral'' is assigned.

We chose to employ the state-of-the-art RTE model proposed by \citet{PNI18} which is a re-implementation of the widely used decomposable attention model developed by \citet{PTDU16}. The model achieves an accuracy of 86.4\% on the SNLI test set.
We selected it because at the time of development of this work, it was one of the best performing systems on the task with publicly available code. Additionally, the usage of preprocessing parsing tools is not required and the model is faster to train when compared to the other approaches we tried. 

Although the model achieved good scores on the SNLI dataset, we noticed that it does not generalize well when employed to predict the relationships between the candidate claim-evidence pairs present in the FEVER data. In order to improve the generalization capabilities of the RTE model, we decided to fine-tune it using a newly synthesized FEVER SNLI-style dataset~\cite{PJ96}. This was accomplished in two steps: the RTE model was initially trained using the SNLI dataset and then re-trained using the FEVER SNLI-style dataset. 

The FEVER SNLI-style dataset was created using the information present in the FEVER dataset while retaining the format of the SNLI dataset. Let us consider each learning instance in the FEVER dataset of the form $\langle c, l, E \rangle$, where $c$ is the claim, $l \in \{$SUPPORTS, REFUTES, NOT ENOUGH INFO$\}$ is the label and $E$ is the set of evidences.
While constructing the FEVER SNLI-style dataset, we only considered the learning instances labeled as ``SUPPORTS'' or ``REFUTES'' because these were the instances that provided us with evidences. Given such an instance, we proceeded as follows: 
for each evidence $e \in E$, we created an SNLI-style example $\langle c, e \rangle$ labeled as ``Entailment'' if $l=$ ``SUPPORTS'' or ``Contradiction'' if $l=$ ``REFUTES''. If $e$ contained more than one sentence, we made a simplifying assumption and only considered the first sentence of $e$. For each ``Entailment'' or ``Contradiction'' which was added to this dataset, a ``Neutral'' learning instance of the form $\langle c, n \rangle$ was also created. $n$ is a randomly selected sentence present the same document from which $e$ was retrieved. We also ensured that $n$ was not included in any of the other evidences in $E$. Following this procedure, we obtain examples that are similar (retrieved from the same document) but should be labeled differently. 
Thus, we obtained a dataset with the characteristics depicted in Table~\ref{tab:FeverSNLIDataset}. To correct the unbalanced nature of the dataset, we performed random undersampling~\cite{HG09}. 
The fine-tuning had a huge positive impact on the generalization capabilities of the model as shown in Table~\ref{tab:finetuneresults}. Using the fine-tuned model, the aforementioned set of probabilities are finally computed.
\setlength\tabcolsep{5pt}
\begin{table}
\begin{center}
\begin{tabular}{|c|c|c|c|} 
\hline
\textbf{Split} & \textbf{Entail.} & \textbf{Contradiction} & \textbf{Neutral} \\
\hline
Training & 122,892 & 48,825 & 147,588 \\
\hline
Dev & 4,685 & 4,921 & 8,184 \\
\hline
Test & 4,694 & 4,930 & 8,432 \\
\hline
\end{tabular}
\caption{FEVER SNLI-style Dataset split sizes for \textsc{Entailment}, \textsc{Contradiction} and \textsc{Neutral} classes}
\label{tab:FeverSNLIDataset}
\end{center}
\end{table}

\setlength\tabcolsep{3pt}
\begin{table}
\begin{center}
\begin{tabular}{|c|c|c|c|c|} 
\hline
\textbf{Model} & \textbf{Macro} & \textbf{Entail.} & \textbf{Contra.} & \textbf{Neutral}\\
\hline
Vanilla & 0.45 & 0.54 & 0.44 & 0.37 \\
\hline
Fine-tuned & 0.70 & 0.70 & 0.64 & 0.77 \\
\hline
\end{tabular}
\caption{Macro and class-specific F1 scores achieved on the FEVER SNLI-style test set}
\label{tab:finetuneresults}
\end{center}
\end{table}
\subsection{Final Classification}
\label{sec:final}
Twelve features were derived using the probabilities computed by the RTE module. 
We define the following variables for notational convenience:
\par
$cs_i = \begin{cases}
            1 & \text{if}\: s_i \geq r_i \:\text{and}\: s_i \geq u_i \\
            0 & \text{otherwise}
\end{cases}$
\par
$cr_i = \begin{cases}
            1 & \text{if}\: r_i \geq s_i \:\text{and}\: r_i \geq u_i \\
            0 & \text{otherwise}
\end{cases}$
\par
$cu_i = \begin{cases}
            1 & \text{if}\: u_i \geq s_i \:\text{and}\: u_i \geq r_i \\
            0 & \text{otherwise}
\end{cases}$
\par


The twelve features which were computed are:
\renewcommand\labelitemi{}
\setlength{\columnsep}{-17pt}
\begin{multicols}{2}
\begin{itemize}
    \item $f_1 = \sum_{i=1}^{n} cs_i$
    \item $f_2 = \sum_{i=1}^{n} cr_i$
    \item $f_3 = \sum_{i=1}^{n} cu_i$
    \item $f_4 = \sum_{i=1}^{n} (s_i \times cs_i)$
    \item $f_5 = \sum_{i=1}^{n} (r_i \times cr_i)$
    \item $f_6 = \sum_{i=1}^{n} (u_i \times cu_i)$
    \item $f_7 = max(s_i) \: \forall i$
    \item $f_8 = max(r_i) \: \forall i$
    \item $f_9 = max(u_i) \: \forall i$
    \item $f_{10} =  
    \begin{cases}
    \frac{f_4}{f_1} \: \text{if} \: f_1 \neq 0 \\
    0 \; \text{otherwise}
    \end{cases}
    $
    \item $f_{11} =  
    \begin{cases}
    \frac{f_5}{f_2} \: \text{if} \: f_2 \neq 0 \\
    0 \; \text{otherwise}
    \end{cases}
    $
    \item $f_{12} =  
    \begin{cases}
    \frac{f_6}{f_3} \: \text{if} \: f_3 \neq 0 \\
    0 \; \text{otherwise}
    \end{cases}
    $
\end{itemize}
\end{multicols}

Each of the possible evidential sentences supports a certain label more than the other labels (this can be determined by looking at the computed probabilities). The variables $cs_i, cr_i$ and $cu_i$ are used to capture this fact. The most obvious way to label a claim would be to assign the label with the highest support to the claim. Hence, we chose to use the features $f_1, f_2$ and $f_3$ which represent the number of possible evidential sentences which support each label. The amount of support lent to a certain label by supporting sentences could also be useful in performing the labelling. This motivated us to use the features $f_4, f_5$ and $f_6$ which quantify the amount of support for each label. If a certain sentence can strongly support a label, it might be prudent to assign that label to the claim. Hence, we use the features $f_7, f_8$ and $f_9$ which capture how strongly a single sentence can support the claim. Finally, we used the features $f_{10}, f_{11}$ and $f_{12}$ because the average strength of the support lent by supporting sentences to a given label could also help the classifier.

These features were used by a Random Forest classifier~\cite{Breiman:2001:RF:570181.570182} to determine the label to be assigned to the claim. The classifier was composed of 50 decision trees and the maximum depth of each tree was limited to 3. Information gain was used to measure the quality of a split. 3000 claims labelled as "SUPPORTS", 3000 claims labelled as "REFUTES" and 4000 claims labelled as "NOT ENOUGH INFO" were randomly sampled from the training set. Relevant sentences were then retrieved as detailed in Section~\ref{sec:docsent_retrieval} and supplied to the RTE module (Section~\ref{sec:rte}). The probabilities calculated by this module were used to generate the aforementioned features. The classifier was then trained using these features and the actual labels of the claims.

We used the trained classifier to label the claims in the test set. If the "SUPPORTS" label was assigned to the claim, the five documents with the highest $(s_i \times cs_i)$ products were returned as evidences. However, if $cs_i = 0 \: \forall i$, then the label was changed to "NOT ENOUGH INFO" and a null set was returned as evidence. A similar process was employed when the "REFUTES" label was assigned to a claim. 
If the "NOT ENOUGH INFO" label was assigned, a null set was returned as evidence.

\section{Results and Discussion}
\label{results}
\setlength\tabcolsep{3pt}
\begin{table}
\begin{center}
\begin{tabular}{|c|c|c|c|} 
\hline
\textbf{Metric} & \textit{DeFactoNLP} & \textit{Baseline} & \textit{Best} \\
\hline
Label Accuracy & 0.5136 & 0.4884 & 0.6821 \\
\hline
Evidence F1 & 0.4277 & 0.1826 & 0.6485 \\
\hline
FEVER Score & 0.3833 & 0.2745 & 0.6421 \\
\hline
\end{tabular}
\caption{System Performance}
\label{perf}
\end{center}
\end{table}

Our system was evaluated using a blind test set which contained 19,998 claims. Table~\ref{perf} compares the performance of our system with that of the baseline system. It also lists the best performance for each metric. The evidence precision of our system was 0.5191 and its evidence recall was 0.3636. All of these results were obtained upon submitting our predictions to an online evaluator. DeFactoNLP had the $5^{th}$ best evidence F1 score, the $11^{th}$ best label accuracy and the $12^{th}$ best FEVER score out of the 24 participating systems.

\par
The results show that the evidence F1 score of our system is much better than that of the baseline system. However, the label accuracy of our system is only marginally better than that of the baseline, suggesting that our final classifier is not very reliable. The low label accuracy may have negatively affected the other scores. Our system's low evidence recall can be attributed to the primitive methods employed to retrieve the candidate documents and sentences. 
Additionally, the RTE module can only detect entailment between two pairs of sentences. 
Hence, claims which require more than one sentence to verify them cannot be easily labelled by our system.
This is another reason behind our low evidence recall, FEVER score and label accuracy. 
We aim to study more sophisticated ways to combine the information obtained from the RTE module in the near future.
%

To better assess the performance of the system,
we performed a manual analysis of the predictions made by the system.
We observed that for some simple claims (\textit{ex.}``Tilda Swinton is a vegan'') which were labeled as ``NOT ENOUGH INFO'' in the gold-standard, the sentence retrieval module found many sentences related to the NEs in the claim but none of them had any useful information regarding the claim object (\textit{ex.}``vegan''). In some of these cases, the RTE module would label certain sentences as either supporting or refuting the claim, even if they were not relevant to the claim. 
In the future, we aim to address this shortcoming by exploring triple extraction-based methods to weed out certain sentences~\cite{gerber2015}.

We also noticed that the usage of coreference in the Wikipedia articles was responsible for the system missing some evidences as the RTE module could not accurately assess the sentences which used coreference. Employing a coreference resolution system at the article level is a promising direction to address this problem.

The incorporation of named entity disambiguation into the sentence and document retrieval modules could also boost performance. This is because we noticed that in some cases, the system used information from unrelated Wikipedia pages whose names were similar to those of the NEs mentioned in a claim to incorrectly label it
(\textit{ex.} a claim was related to the movie ``Soul Food'' but some of the retrieved evidences were from the Wikipedia page related to the soundtrack ``Soul Food'').

\section{Conclusion}
\label{conclusion}
In this work, we described our fact verification system, DeFactoNLP, which was designed for the FEVER 2018 Shared Task. When supplied a claim, it makes use of NER and TFIDF vector comparison to retrieve candidate Wikipedia sentences which might help in the verification process. An RTE module and a Random Forest classifier are then used to determine the veracity of the claim based on the information present in these sentences. 
The proposed system achieved a 0.4277 evidence F1-score, a 0.5136 label accuracy and a 0.3833 FEVER score.
After analyzing our results, we have identified many ways of improving the system in the future. For instance, triple extraction-based methods can be used to improve the sentence retrieval component as well as to improve the identification of evidential sentences. We also wish to explore more sophisticated methods to combine the information obtained from the RTE module and employ entity linking methods to perform named entity disambiguation.

\section*{Acknowledgments}
This research was partially supported by an EU H2020 grant provided for the WDAqua project (GA no. 642795) and by the DAAD under the ``International promovieren in Deutschland – für alle'' (IPID4all) project.

\bibliography{emnlp2018}
\bibliographystyle{acl_natbib_nourl}

\end{document}